\documentclass{article}

\usepackage{amsmath,amsfonts,bm}

\def\eqref#1{equation~\ref{#1}}

\def\1{\bm{1}}

\def\rmW{{\mathbf{W}}}

\def\vb{{\bm{b}}}
\def\vc{{\bm{c}}}

\def\vh{{\bm{h}}}

\def\vs{{\bm{s}}}

\def\mF{{\bm{F}}}

\def\mI{{\bm{I}}}

\def\mM{{\bm{M}}}

\def\mT{{\bm{T}}}

\DeclareMathAlphabet{\mathsfit}{\encodingdefault}{\sfdefault}{m}{sl}
\SetMathAlphabet{\mathsfit}{bold}{\encodingdefault}{\sfdefault}{bx}{n}

\def\gJ{{\mathcal{J}}}

\newcommand{\E}{\mathbb{E}}

\newcommand{\R}{\mathbb{R}}

\usepackage{microtype}
\usepackage[]{graphicx}
\usepackage{booktabs} 
\usepackage{hyperref}
\usepackage{url}
\usepackage{adjustbox}
\usepackage[toc,page]{appendix}
\usepackage{subcaption}
\usepackage[table]{xcolor}

\usepackage[accepted]{icml2020}

\def\Wcore{{\rmW_\textrm{core}}}
\def\Wfac{{\rmW_\textrm{fac}}}
\def\Wmem{{\rmW_\textrm{mem}}}
\def\Wm{{\rmW_\textrm{M}}}
\def\Wac{{\rmW_\textrm{ac}}}
\def\Wpi{{\rmW_\pi}}
\def\Wv{{\rmW_\textrm{V}}}

\icmltitlerunning{Working Memory Graphs}

\begin{document}

\twocolumn[
\icmltitle{Working Memory Graphs}



\icmlsetsymbol{equal}{*}

\begin{icmlauthorlist}
\icmlauthor{Ricky Loynd}{to}
\icmlauthor{Roland Fernandez}{to}
\icmlauthor{Asli Celikyilmaz}{to}
\icmlauthor{Adith Swaminathan}{to}
\icmlauthor{Matthew Hausknecht}{to}
\end{icmlauthorlist}

\icmlaffiliation{to}{Microsoft Research AI, Redmond, Washington, USA}

\icmlcorrespondingauthor{Ricky Loynd}{riloynd@microsoft.com}

\icmlkeywords{Reinforcement Learning, Transformers}

\vskip 0.3in
]

\printAffiliationsAndNotice{}

\begin{abstract}
Transformers have increasingly outperformed gated RNNs in obtaining new state-of-the-art results on supervised tasks involving text sequences. 
Inspired by this trend, we study the question of how Transformer-based models can improve the performance of sequential decision-making agents.
We present the \textit{Working Memory Graph} (WMG), an agent that employs multi-head self-attention to reason over a dynamic set of vectors representing observed and recurrent state.
We evaluate WMG in three environments featuring factored observation spaces: a Pathfinding environment that requires complex reasoning over past observations, BabyAI gridworld levels that involve variable goals, and Sokoban which emphasizes future planning.
We find that the combination of WMG's Transformer-based architecture with factored observation spaces leads to significant gains in learning efficiency compared to baseline architectures across all tasks.
WMG demonstrates how Transformer-based models can dramatically boost sample efficiency in RL environments for which observations can be factored.
\end{abstract}

\section{Introduction}

Because of their ability to process sequences of data, gated Recurrent Neural Networks (RNNs) have been widely applied to natural language processing (NLP) tasks such as machine translation. In the RNN-based approach of \citet{Sutskever:2014:SSL:2969033.2969173}, an encoder RNN maps an input sentence in the source language to a series of internal hidden state vectors. The encoder's final hidden state is copied into a decoder RNN, which then generates another sequence of hidden states that determine the selection of output tokens in the target language. This model can be trained to translate sentences, but translation quality deteriorates on long sentences where long-term dependencies become critical. 
A plausible conjecture attributes this drop in performance to the limited representational capacity of RNN hidden state vectors.
In \citet{bahdanau2014neural}, translation quality is boosted by applying an attention mechanism to create paths serving as shortcuts from the input to the output sequences, routing information outside the linear chain of the RNN's hidden states.
Similar attention mechanisms have since gained wide usage, culminating in the Transformer model \citep{DBLP:journals/corr/VaswaniSPUJGKP17} which replaces the RNN with many short paths of self-attention. Since then, Transformers have outperformed RNNs on many NLP tasks \citep{DBLP:journals/corr/abs-1810-04805,unilmmsr,roberta},
and have been successfully applied to set-structured data \citep{SetTransformer, PointClouds}.

\begin{figure*}[htp]
    \begin{subfigure}{0.22\textwidth}
    {
    \centering
    \includegraphics[width=0.95\textwidth]{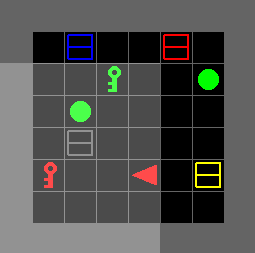}
    \begin{center}
    \small
    \textit{\textbf{Pick up the green key}}
    \end{center}
    }
    \end{subfigure}
    \begin{subfigure}{0.77\textwidth}
    {
    \centering
    \includegraphics[width=\linewidth]{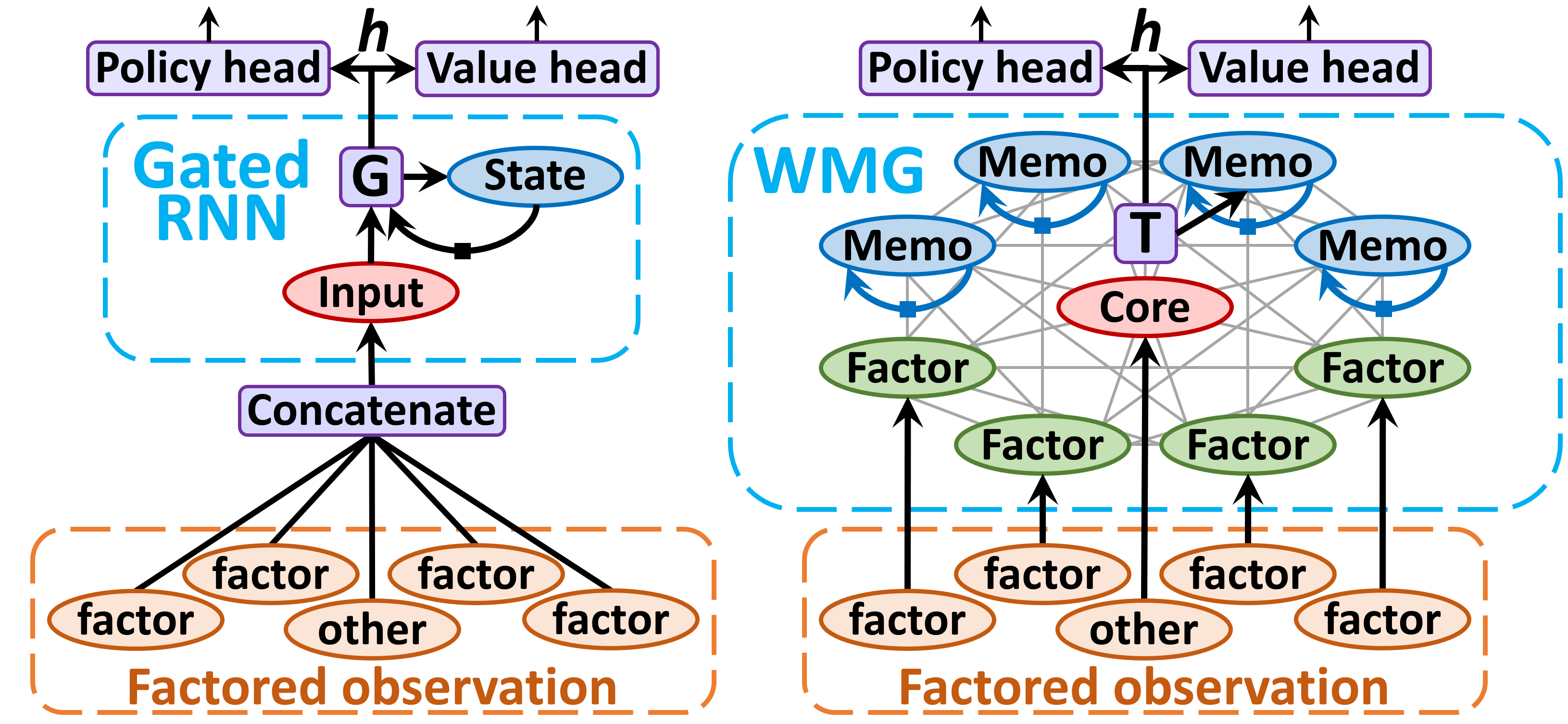}
    }
    \end{subfigure}
    \caption{
    \textbf{Left:} BabyAI rewards the agent (red triangle) for performing the task given by the instruction. Here, the agent looks to the left. Its 7x7 field of view (shown in lighter grey) extends beyond the walls in this case. \textbf{Right:} Comparing a gated RNN agent to WMG: a gated RNN constrains information to flow through a long, time-sequential path of hidden state vectors, whereas WMG allows information to flow through shorter, parallel paths of self-attention among its set of \emph{Memo} vectors. \emph{Factor} vectors embed observation factors and are used along with Memos and a Core vector as input to WMG's multi-layer Transformer \textit{T}, which effectively replaces the gating layers of a gated RNN.
}
    \label{babyai_intro_example}
\end{figure*}

We seek to leverage these intuitions to improve the ability of Reinforcement Learning (RL) agents to reason over long time horizons in Partially Observable Markov Decision Processes (POMDPs)~\citep{kaelbling98}.
In a POMDP, a single observation $Obs_t$ is not sufficient to identify the latent environment state $s_t$.
Thus the agent must reason over the history of past observations in order to select the best action for the current timestep.
A simple strategy employed by DQN \citep{mnih2015humanlevel} is to condition the policy on the $N$ most recent observations $\pi(a_t|Obs_{t-N+1} \dots Obs_{t})$. 
But in complex environments, the sufficient number $N$ may be large, highly variable, and unknown. 
To address this issue, gated RNNs such as LSTMs \citep{Hochreiter:1997:LSM:1246443.1246450} and GRUs \citep{GRU, Chung:2015:RLV:2969442.2969572} use internal, recurrent state vectors which can in theory maintain information from past observations \citep{DBLP:journals/corr/HausknechtS15, Oh:2016:CMA:3045390.3045684}.
In practice however, these methods are limited by the single path of information flow defined by the linear chain of RNN hidden states.
As in NLP, we hypothesize that providing alternative paths for information to follow will be advantageous to RL agents.
Building on this intuition, we introduce the Working Memory Graph (WMG), a Transformer-based agent that uses self-attention to provide a multitude of shortcut paths for information to flow from past observations to the current action
through a dynamic set of hidden state vectors called \emph{Memos}, illustrated in Figure~\ref{babyai_intro_example}~(right).

Motivated by prior work on factored representations~\citep{Russell:2009:AIM:1671238} and factored MDPs \citep{Boutilier:2000:DHA:647288.721273, Boutilier:2001:SDP:1642090.1642184}, we argue that factored observations are ideally suited for processing by Transformer-based agents like WMG.
Although many environments use fixed-sized feature spaces, many other domains have observations amenable to factoring.
As a motivating example, consider the BabyAI environment ~\citep{Chevalier-2018} as depicted in Figure~\ref{babyai_intro_example} (left). 
The native observation space includes the agent's field of view, a 7x7 region in front of it.
This observation can be efficiently represented by a set of factors describing the types, colors, and relative $x$ and $y$ coordinates of all objects currently in the field of view: {\textbf{([green, key, 3, 1], [grey, box, 1, 2], [green, ball, 2, 2], [red, key, 0, 3])}}.
This \textit{factored observation} is more compact than the native observation, but will vary in size depending on the number of objects in view.
Each observation factor (e.g. \textbf{[grey, box, 1, 2]}) is embedded into a \emph{Factor vector} (Fig. \ref{babyai_intro_example} right) which serves as input to WMG's Transformer, along with other Factors and the Memos.

\textbf{Our contributions are twofold}: First we introduce the \textit{Working Memory Graph} (WMG), a Transformer-based agent implementing a novel form of \textit{shortcut recurrence} which we demonstrate to be effective at complex reasoning over long-term dependencies.
Second, we identify the synergy between Transformer-based RL architectures and \textit{factored observations}, demonstrating that by virtue of self-attention, WMG is able to effectively leverage factored observations to learn high-performing policies using fewer environment interactions than alternative architectures.

\section{Related Approaches}
Prior approaches for reasoning over long time horizons use attention for memory access \citep{graves2016hybrid, Oh:2016:CMA:3045390.3045684} or self-attention to process individual observations \citep{HkxaFoC9KQ, vinyals2019grandmaster, RTFM}. These approaches rely on LSTM-based recurrence over sequences.
In contrast, WMG obviates the need for gated recurrence by applying self-attention to a network of Memos persisted through time.

After attempting to apply Transformer-style attention to RL tasks, \citet{SNAIL} concluded that such architectures could not easily process sequential information. 
Other models handle partial observability using gated RNNs with Transformer-style self-attention over state vectors analogous to WMG's Memos, but with different state-update schedules: 
RMC \citep{DBLP:journals/corr/abs-1806-01822} updates \textit{all} state vectors on every timestep, while RIMs \citep{DBLP:journals/corr/abs-1909-10893} enforces sparsity by updating exactly \textit{half} of the state vectors 
on each step.
In contrast, WMG replaces only \textit{one} Memo on each timestep to maximize Memo persistence and facilitate preservation of information through time. 

Unlike the other models discussed here, the Gated Transformer-XL \citep{GTrXL} addresses partial observability by feeding hundreds of past observations into the Transformer at once. 
In order to mitigate the Transformer's $O(N^2)$ computational cost in number of inputs, WMG instead computes self-attention over a much smaller number of recurrent Memos to capture and maintain relevant aspects of past observations.
Another significant difference is that GTrXL relies on inserting \textit{gated RNNs} into its Transformer, while WMG applies the original \textit{unmodified} Transformer design to RL.

\section{Working Memory Graph}
\begin{figure*}[t!]
\begin{center}
\includegraphics[width=1.0\linewidth]{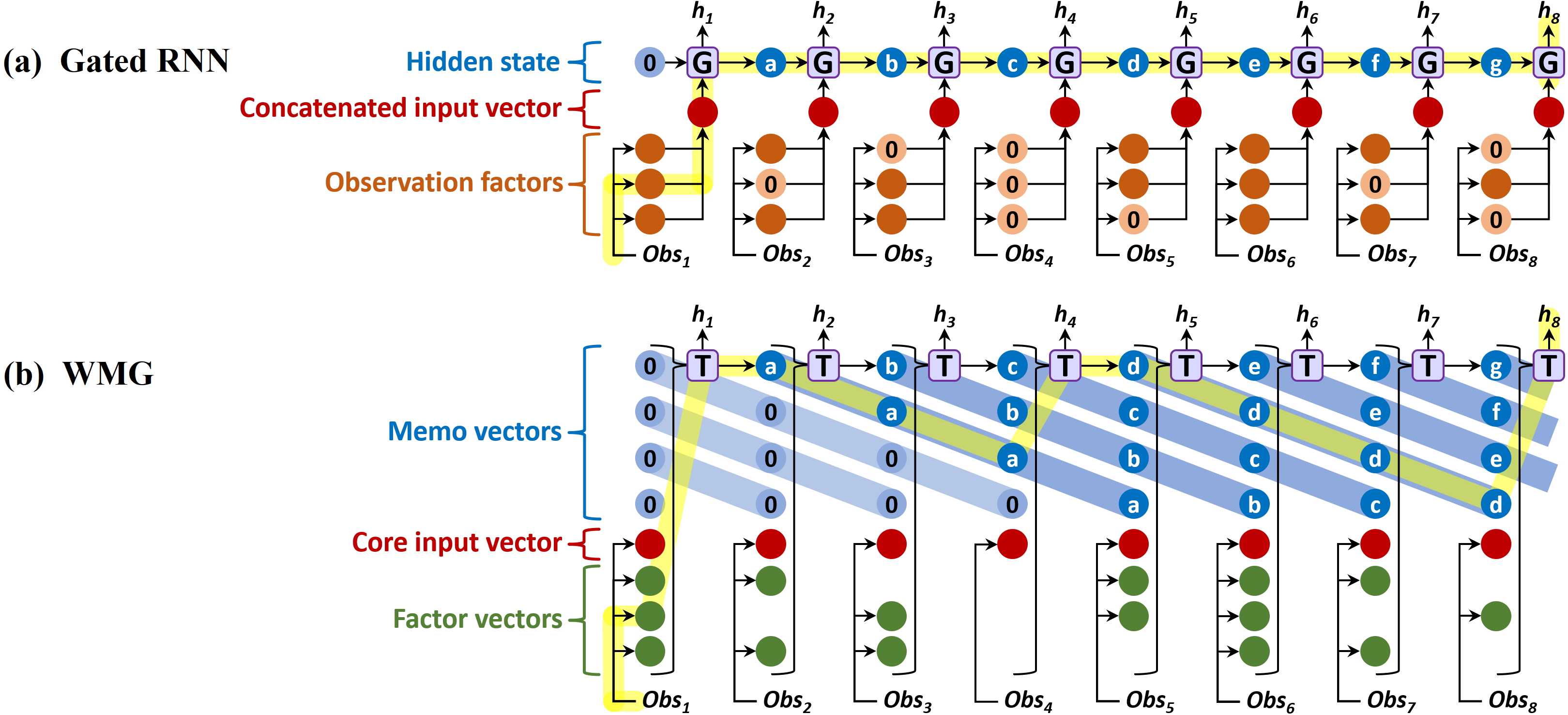}
\caption{
\textbf{Unrolled Over Time}:
In a \textbf{Gated RNN} (a), in order for the first observation $Obs_1$ to affect the agent's output $\vh_8$, information must pass through 8 gating operations and 7 intervening hidden states \textbf{a-g}. In a \textbf{WMG} (b), many possible paths lead from $Obs_1$ to $\vh_8$. The highlighted path requires only three passes through \textbf{T}, as the information is first stored for several timesteps in Memo \textbf{a} and later in Memo \textbf{d}. 
Information flows forward, and gradients flow backward, over many such shortcut paths.}
\label{unrolled_fig}
\end{center}
\end{figure*}

The term \textit{Working Memory Graph} is motivated by the \textit{limited size} of WMG's self-attention computation graph, in loose analogy with the cognitive science term \textit{working memory}, referring to a cognitive system that holds a limited amount of information for use in mental processing \citep{mil56}.
As illustrated in Figure~\ref{babyai_intro_example}, WMG applies multi-head self-attention to a dynamic set of hidden state vectors, called \emph{Memos}, which store information from previous timesteps.
Formally, each Memo vector defines one row in a Memo matrix $\displaystyle \mM \in \R^{n_M {\times} d_M}$, where $n_M$ is the number of Memos maintained by WMG and $d_M$ is the size of each Memo vector. 
Any Memos present at the start of an episode are initialized to zero.
As a rolling buffer, the matrix persists each Memo unchanged through $n_M$ timesteps.
For example, in Figure~\ref{unrolled_fig}~(b) the Memo~$b$ is created on the third step and persists unchanged for 4 steps before being overwritten by Memo~$f$. 
Memos are the basis of WMG's \textit{shortcut recurrence}, replacing a gated RNN's single path of information flow with a network of shorter self-attention paths. 

In addition to Memos, WMG also applies self-attention to a variable number of \emph{Factor} vectors derived from observations, depicted in green in Figures~\ref{babyai_intro_example},~\ref{unrolled_fig}.
On each timestep, WMG receives an observation consisting of a variable number of factors, which are copied into ($n_F$) Factor vectors forming a Factor matrix $\displaystyle \mF \in \R^{n_F {\times} d_F}$.
Finally, a single Core vector $\displaystyle \vc \in \R^{d_c}$ encodes any non-factored portions of the observation. 
The Core, Factors and Memos are embedded and stacked to form the Transformer input matrix:
\[
\displaystyle 
  \mT^{in}=
  \begin{bmatrix}
    \vc \Wcore + \vb_\textrm{core} \\
    \mF \Wfac + \vb_\textrm{fac} \\
    \mM'\Wmem + \vb_\textrm{mem}
  \end{bmatrix},\;\;\;\;
  \mM'=[\mM \;\; \mI]
\]

where $\displaystyle \Wcore \in \R^{d_c {\times} d_T}$,
$\displaystyle \Wfac \in \R^{d_F {\times} d_T}$ and
$\displaystyle \Wmem \in \R^{(d_M+n_M) {\times} d_T}$
are embedding matrices with corresponding bias vectors $\displaystyle \vb \in \R^{d_T}$ broadcast over rows. 
Each Memo is concatenated with a one-hot vector (from the identity matrix) indicating its age.

Closely following the encoder architecture of  \citet{DBLP:journals/corr/VaswaniSPUJGKP17}, WMG's Transformer takes the input matrix $\displaystyle \mT^{in} \in \R^{n_T {\times} d_T}$ and returns an output matrix $\displaystyle \mT^{out} \in \R^{n_T {\times} d_T}$, where $\displaystyle n_T=1+n_F+n_M$ is the number of input (or output) vectors, and $\displaystyle d_T$ is the size of each vector. 
On each timestep, the oldest Memo (final row of the Memo matrix) is replaced by a new Memo (as the incoming first row) generated by a non-linear function of the Core's output vector $\displaystyle \vh = \mT_{0:}^{out}$:
\[
\begin{split}
\displaystyle \mM=
  \begin{bmatrix}
    \mathit{tanh}(\vh \Wm + \vb_{M}) \\
    \mM_{0:-1}
  \end{bmatrix} \\
  \Wm \in \R^{d_T {\times} d_M},\;
  \vb_M \in \R^{d_M}
\end{split}
\]

The actor-critic network applies a shared linear layer to $\vh$, followed by $\mathit{ReLU}$ activation to obtain $\vs_\textrm{ac}$. This is followed by two separate linear layers to produce the agent's policy and value outputs $\pi$ and $V$:
\[
\begin{split}
\displaystyle 
  \vs_\textrm{ac} = \mathit{ReLU}(\vh \Wac + \vb_\textrm{ac}),\;
  \Wac \in \R^{d_T {\times} d_{ac}},\;
  \vb_\textrm{ac} \in \R^{d_{ac}} \\
  \pi = \mathit{softmax}(\vs_\textrm{ac} \Wpi + \vb_\pi),\;
  \Wpi \in \R^{d_{ac} {\times} d_\pi},\;
  \vb_\pi \in \R^{d_\pi} \\
  V = \vs_\textrm{ac} \Wv + b_V,\;
  \Wv \in \R^{d_{ac}} \\
\end{split}
\]

\textbf{RL Training.}
To summarize, the trainable parameters $\theta$ of WMG include all Transformer parameters, plus (for embeddings) $\Wcore, \vb_\textrm{core}, \Wfac, \vb_\textrm{fac}, \Wmem, \vb_\textrm{mem}$, (for Memo creation) $\Wm, \vb_\textrm{M}$, (actor-critic shared) $\Wac, \vb_\textrm{ac}$, (actor) $\Wpi, \vb_\pi$, and (critic) $\Wv, b_V$. 
Parameters $\theta$ are trained end-to-end through backpropagation of the standard actor-critic, policy-gradient loss functions, maximizing an entropy-regularized expected return of the actor, and minimizing a k-step TD error of the critic. 
The entropy-regularized policy gradient is:
\[
\begin{split}
\nabla_\theta \gJ(\theta) = \E_\pi [ \sum_{t=0}^\infty \nabla_\theta \log \pi(a_t|\vh_t;\theta) A^\pi(Obs_t,a_t) \\
+ \beta \nabla_\theta H(\pi(\vh_t;\theta))]
\end{split}
\]
where $\displaystyle \pi(a|\vh_t;\theta)$ denotes WMG's policy head operating on hidden state $\vh_t$ (see Fig.~\ref{babyai_intro_example} right), $H$ is the entropy of the policy's action distribution, and $\beta$ controls the strength of the entropy regularization term. When performing backpropagation through time, the maximum number of steps of gradient flow is denoted by $t_{max}$. To reduce the variance of gradient estimates, we use the A3C algorithm described by \citet{Mnih16Asynchronous}, which estimates the advantage $A^\pi(s_t,a_t)$ using a $\gamma$-discounted $k$-step return as follows:
\[
A^\pi(Obs_t,a_t) = (\sum_{i=0}^{k-1} \gamma^i r_{t+i}) + \gamma^k V(\vh_{t+k}) - V(\vh_t;\theta)
\]
where $V(\vh_t;\theta)$ denotes WMG's state-value head (see Fig.~\ref{babyai_intro_example} right), which is trained to minimize the squared difference between the $k$-step return and the current value estimate $|| (\sum_{i=0}^{k-1} \gamma^i r_{t+i} + \gamma^k V(\vh_{t+k})) - V(\vh_t) ||^2$, and k is upper-bounded by the number of timesteps ($t_{max}$) in the actor's current update window.

To encourage further work and comparative studies, we provide WMG's source code and pre-trained models at \url{https://github.com/microsoft/wmg_agent}. 

\section{Experiments}

Our experiments aim to (1) evaluate WMG's ability to reason over long time spans in a setting of high partial observability, and (2) understand how factored representations may be effectively utilized by WMG.
To address these questions we present results on three diverse environments: a novel Pathfinding task which requires complex reasoning over past observations, the BabyAI domain \citep{Chevalier-2018} which involves changing goals, partial observability, and textual instructions, and Sokoban~\citep{Sokoban}, a challenging puzzle environment that benefits from forward planning ability.
To foreshadow our results, the Pathfinding task establishes the effectiveness of WMG's shortcut recurrence, BabyAI demonstrates that WMG leverages factored observations to deliver substantial gains in sample efficiency, while Sokoban shows that WMG can learn to solve very difficult tasks. 
These experimental results illustrate WMG's ability to handle highly diverse and demanding environments for which observations can be factored.

In all experiments, we conduct extensive hyperparameter tuning of each agent (including the baselines) using a guided form of random search that we call \textit{Distributed Grid Descent} (DGD). It is designed to address the challenges posed by large numbers of hyperparameters (10-20), and the high variance among independent training runs given the same hyperparameter configuration that is often observed in Deep RL experiments. (See Appendix~\ref{appendix:hyperparameter_tuning} for a detailed description of DGD.) After hyperparameter tuning, in order to remove selection bias, we perform many independent training runs using the tuned values, then report the means and standard deviations over those final training runs.

\begin{figure*}[htp]
    \centering
    \begin{subfigure}{0.30\textwidth}
    \centering
    \includegraphics[width=\linewidth]{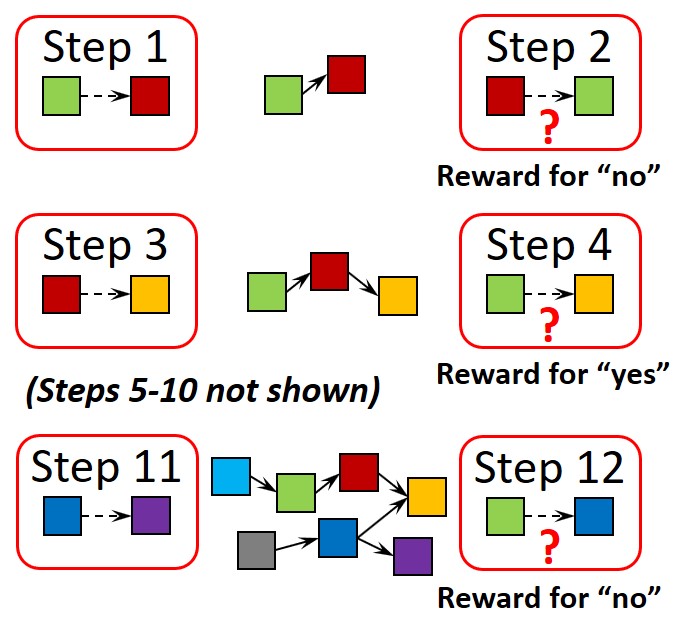}
    \end{subfigure}
    \begin{subfigure}{0.357\textwidth}
    \centering
    \includegraphics[width=\linewidth]{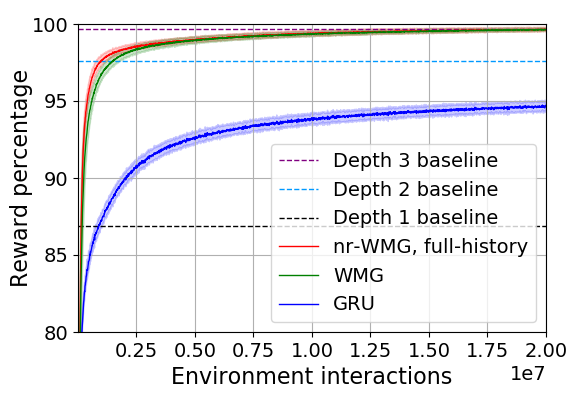}
    \end{subfigure}
    \begin{subfigure}{0.333\textwidth}
    \includegraphics[width=\linewidth]{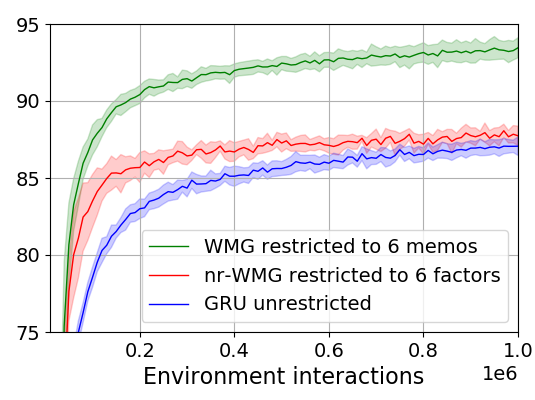}
    \end{subfigure}
    \vspace{-.5em}
    \caption{\textbf{Left: Pathfinding task}, consists of 12 timesteps with a maximum graph size $N=7$ pattern nodes. The boxes with rounded corners illustrate the observations for the given timesteps, where a question mark identifies the step as a quiz step rather than a construction step. The box colors represent distinct pattern vectors which never repeat between episodes. \textbf{Middle: Results on Pathfinding.} 
    Each plotted point is the percentage of reward on quiz steps received by the agent over the previous 10k timesteps, averaged over 100 independent training runs. Bands display one standard deviation. (See Table~\ref{pathfinding_detailed_results} for more details.) \textbf{Right: Probing shortcut recurrence.} WMG restricted to only 6 Memos outperformed both nr-WMG (given access to the last 6 observations) and GRU.}
    \label{pathfinding_fig}
\end{figure*}

\subsection{Pathfinding Task}

Pathfinding is designed to evaluate an agent's ability to perform complex reasoning over past observations.
Figure~\ref{pathfinding_fig}~(left) depicts the incremental construction of a directed acyclic graph over nodes identified by unique pattern vectors which are randomly regenerated on every episode. (See Appendix~\ref{appendix:pathfinding-environment-details} for the graph-construction algorithm and other details.)
On odd timesteps the agent observes the pattern vectors of two nodes to be linked, and on even steps the agent must determine whether or not a directed path exists from one given pattern to the other.
As this cycle repeats, the graph grows larger and the agent must perform an increasing number of reasoning steps to confirm or deny the existence of a path between arbitrary nodes.
Because each observation reveals only part of the graph, the agent must utilize information from previous observations to infer graph connectivity.

For example, consider step 4 of Figure~\ref{pathfinding_fig}~(left): To determine whether a path exists from green to yellow, the agent must recall and combine information from steps 1 and 3. Similarly, on step 12, if the agent were asked about the existence of a path from cyan to yellow, answering correctly without guessing would require piecing together information from three non-contiguous timesteps. Since the actual quiz on step 12 asks whether a path exists from green to blue, the agent must reason over many past observations to determine that no such path exists.

Each pattern is a vector of D real numbers drawn randomly from the interval -1 to 1 on each episode.
A binary value is added to the observation vector to indicate whether the current step is a quiz or construction step, bringing the size of the observation space to $2D+1$, where $D=7$ for our experiments. 
The action space consists of two actions, signifying \textit{yes} and \textit{no}.
The agent receives a reward of 1 for answering correctly on a quiz step.
The quiz questions are constructed to guarantee that each answer (\textit{yes} or \textit{no}) is correct half the time on average, so agents that act randomly or have no memory of past observations will obtain 50\% of possible reward in expectation.

For this task, WMG is configured with Memos but no Factors.
The number of Memos is a tuned hyperparameter, equal to 16 in this experiment.
(See Appendix~\ref{appendix:appendix_tables} for all settings.)
On each timestep, the observation is passed to WMG's Core, 
and WMG generates a new Memo and action distribution.
We compare WMG's performance to several baselines.
Each \textit{Depth-n} baseline is a hand-coded algorithm demonstrating the performance obtained using perfect memory of past observations and perfect reasoning over paths up to $n$ steps long.
For example, \textit{Depth-2} remembers all previous construction steps, and reasons over all paths of depth 2.
Finally, in order to understand the effectiveness of Memos at capturing past information, we evaluate a \textit{full-history}, non-recurrent version of WMG (nr-WMG) by removing the Memos and giving it 
all past observations as separate Factors on each timestep.

As shown in Figure~\ref{pathfinding_fig} (middle), the GRU-based agent exceeded \textit{Depth-1} performance, but remained well short of \textit{Depth-2} performance after 20 million steps of training (environment interactions). 
In contrast, both versions of the WMG agent nearly reached \textit{Depth-3} performance, demonstrating a greater ability to perform complex reasoning over past observations. 
The best performance was achieved by nr-WMG, for which the environment is fully observed. 
But WMG with Memos was nearly as sample efficient as this perfect-memory baseline.
These results indicate that shortcut recurrence enables WMG to learn to store and utilize essential information from past Pathfinding observations in a more effective manner than a GRU's gated recurrence.

To assess zero-shot generalization beyond the horizon of the original Pathfinding task, 
we evaluated these 300 models (100 per agent architecture, trained on 12-step episodes) on 1000 fixed episodes of \textit{24 steps} each. 
With no further training, nr-WMG and WMG respectively obtained 95.3\% and 93.9\% of possible score versus 84.4\% for GRU,
showing significantly better generalization to larger graphs than those seen during training.

To further investigate WMG's shortcut recurrence, we repeated the Pathfinding experiment while restricting the total number of WMG Memos to 6, enough to store only half of an episode's observations. 
As shown in Figure~\ref{pathfinding_fig} (right), while WMG's performance degraded slightly, it significantly outperformed nr-WMG (given the last 6 observations) as well as GRU. nr-WMG with 6 Factors captures the heuristic employed by DQN~\cite{mnih2015humanlevel} of stacking several previous observations to combat partial observability.
This result suggests that WMG leverages shortcut recurrence to transfer information from older Memos to newer Memos in order to reason beyond the last 6 observations more effectively than a GRU with its linear chain of gated recurrence.

\begin{figure*}[htp]
    \begin{subfigure}{0.35\textwidth}
    {
    \centering
    \includegraphics[width=.6\linewidth]{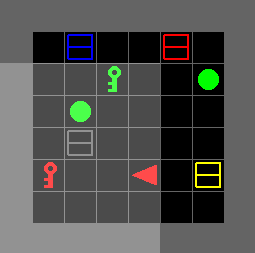}
    \begin{center}
    \textit{\textbf{Go to the yellow box behind you}}
    \end{center}
    }
    \end{subfigure}\;\;
    \begin{subfigure}{0.64\textwidth}
    {
    \small
    \begin{tabular}{l | l | l}
    \textbf{Part of observation} & \textbf{Variable assignments} & \textbf{Vector} \\ 
    \hline
    Factored image & color=\textbf{green}, type=\textbf{key}, X=\textbf{3}, Y=\textbf{1} & Factor \\
    \hline
    Factored image & color=\textbf{grey}, type=\textbf{box}, X=\textbf{1}, Y=\textbf{2} & Factor \\
    \hline
    Factored image & color=\textbf{green}, type=\textbf{ball}, X=\textbf{2}, Y=\textbf{2} & Factor \\
    \hline
    Factored image & color=\textbf{red}, type=\textbf{key}, X=\textbf{0}, Y=\textbf{3} & Factor \\
    \hline
    Factored image & vertical wall X=\textbf{-2} & Core \\
    \hline
    Factored image & horizontal wall Y=\textbf{4} & Core \\
    \hline
    Factored instruction & command=\textbf{go to}, article=\textbf{the}, & Core \\
    & color=\textbf{yellow}, type=\textbf{box}, loc=\textbf{behind you} & \\
    \hline
    Additional info & orientation=\textbf{west}, & Core \\
    & last action=\textbf{move forward} & \\
    \hline
    \end{tabular}
        }
    \end{subfigure}
    \caption{One completely factored observation, where each variable assignment corresponds to a one-hot vector. 
    Since the number of objects in an observation can vary, 
    each object's vectors are concatenated then passed to a single Factor.
    All non-factored parts of the observation are concatenated then passed to the Core. 
    X \& Y coordinates refer to a frame of reference with the agent at the origin, pointed in the positive Y direction. The agent always observes one vertical wall and one horizontal wall.}
    \label{factored_observation}
\end{figure*}

\subsection{BabyAI Environment}
In order to understand how factored representations may be effectively utilized by WMG, we study BabyAI, a domain whose observation space is amenable to factoring.
BabyAI \citep{Chevalier-2018} is a partially observable, 2D grid-world containing objects that can be viewed and moved by the agent.
Unlike most RL environments, BabyAI features text instructions that specify the goal, such as ``pick up the green box''.

We focus on five BabyAI levels, for which the environment consists of a single 6x6 room, as shown in Figure~\ref{factored_observation} (left).
Despite the apparent simplicity of a single-room domain, learning to solve it can often take model-free RL agents hundreds of thousands of environment interaction steps  \citep{Chevalier-2018}.
The agent's action space consists of 7 discrete actions:  Move forward, Turn left, Turn right, Pick up, Drop, Toggle, and Done. 
An episode ends after 64 timesteps, or when the agent achieves the goal, for which it receives a reward of 1.
In Level 1 (GoToObj), the room contains only one object. The agent completes the mission by moving to an adjacent square and pointing toward the object. In Level 2, the target object is always a red ball, and seven grey boxes are present as distractors. In Level 3, the distractors may be any of the 3 object types and 6 colors. If one of the distractors happens to be a red ball, the agent is rewarded for reaching it. In Level 4, the instruction specifies the color and type of the target object. This is the first level in which the text instruction contains valuable information. (See Table~\ref{babyailevels} for instruction templates.) Level 5 increases the difficulty of Level 4 in two ways: First, the agent must reach and pick up the target object. Second, if multiple qualifying target objects are present, the agent is given the initial relative location of the true target, such as ``behind you''.

Throughout this work we follow the strategy of routing multi-instance aspects of observations to WMG Factors, and single-instance aspects to the Core.
Figure~\ref{factored_observation} gives the factoring details for BabyAI, 
where each agent observation consists of a text instruction, an image, and the agent's orientation.
The image's native format is a 7x7 array of cell descriptors (not pixels) identifying three attributes of each cell:  type, color, and open/closed/locked (referring to doors, which are not found in these 5 levels).
In our experiments the text instruction is always factored in a fixed style, while the image is formatted in various ways to study agent capabilities: 
(1) \textbf{7x7x3}, the native BabyAI image array; (2) \textbf{flat}, the native 7x7x3 array flattened to one vector; (3) \textbf{factored} image, as described in Figure~\ref{factored_observation}. (\textit{Note} that when a factored image is passed to a GRU, it is first flattened and padded to form a fixed-length vector.)

To determine whether WMG can leverage factored observations more effectively than gated RNNs in BabyAI, we evaluate the following agents:
(1) \textbf{WMG} is the full, recurrent WMG model, with Factors from observations, (2) \textbf{nr-WMG} is an ablated, non-recurrent version of WMG with no Memos, and no access to prior observations, (3) \textbf{GRU}, and (4) \textbf{CNN+GRU} uses a CNN to process the native 7x7x3 image, followed by a GRU. 
This CNN is one of the two CNN models provided in the BabyAI open source code \citep{Chevalier-2018}.

\begin{table*}[t!]
\caption{
\textbf{BabyAI sample efficiency}: the amount of training (shown here in thousands of environment interactions) required for a model to solve 99\% of 10,000 new, random episodes.
Hyperparameters were first tuned on each model/format/level combination separately, 
then each reported result was computed as the median sample efficiency over 100 additional training runs.
Dashes indicate that no model reliably reached a solution rate of 99\% within 6 million training steps (environment interactions).
Note that \citet{Chevalier-2018} report sample efficiencies in terms of episodes rather than environment interactions. (See Table~\ref{babyailevels}.)
}
\label{babyaimodelperformance}
\begin{center}
\begin{tabular}{l | r r r r r r r r}
Model & WMG & nr-WMG & GRU & WMG & GRU & CNN+GRU \\ 
Image format & factored & factored & factored & flat & flat & native 7x7x3 \\ 
\hline
Level 1 - GoToObj & 1.6 & \textbf{1.4} & 1.7 & 15.0 & 19.0 & 10.6 \\
Level 2 - GoToRedBallGrey & 6.7 & \textbf{5.2} & 24.6 & 29.0 & 31.0 & 22.3 \\
Level 3 - GoToRedBall & \textbf{16.0} & 23.6 & 174.4 & 92.0 & 124.6 & 204.9 \\
Level 4 - GoToLocal & \textbf{59.7} & 71.3 & 2,241.6 & 1,379.9 & 1,799.4 & ----- \\
Level 5 - PickupLoc & \textbf{222.3} & 253.0 & ----- & ----- & ----- & ----- \\
\hline
\end{tabular}
\end{center}
\end{table*}

\begin{figure*}[htp]
    \centering
    \begin{subfigure}{0.55\textwidth}
    {
    \begin{center}
        \begin{tabular}{l | r r r r r r r}
        Model & WMG & nr-WMG & GRU & WMG & GRU \\ 
        Image & factored & factored & factored & flat & flat \\ 
        \hline
        Level 1 & 5.0 & \textbf{3.2} & 8.0 & 40.6 & 36.9 \\
        Level 2 & 13.5 & \textbf{9.9} & 42.6 & 74.9 & 55.3 \\
        Level 3 & \textbf{34.7} & 39.3 & 313.9 & 231.4 & 188.9 \\
        \hline
        \end{tabular}
        \end{center}
    }
    \end{subfigure}
    \begin{subfigure}{0.44\textwidth}
    \centering
    \includegraphics[width=.9\linewidth]{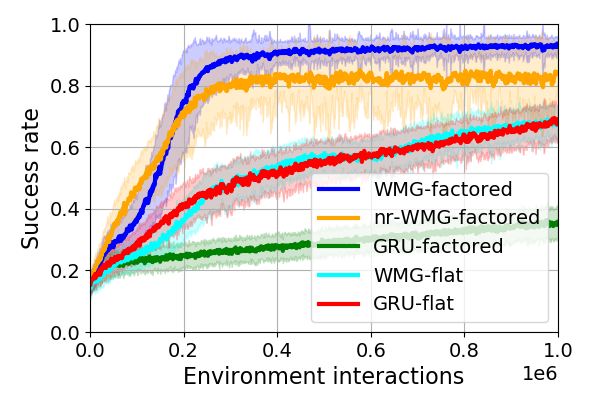}
    \end{subfigure}
    \vspace{-1em}
    \caption{\textbf{Hyperparameter Sensitivity}: \textbf{(Left)} Sample efficiency (in thousands of environment interactions to reach 99\% success rate) of various model-format combinations on Levels 1-3, using hyperparameters optimized for Level 4 (GoToLocal). 
    \textbf{(Right)} Performance on Level 5 (PickupLoc) using hyperparameters optimized for Level 4. 
    Although none of the models reach 99\% success rate on Level 5, WMG with factors reaches a high level of performance before the others.}
    \label{fig:hyperparam_sensitivity}
\end{figure*}

\subsubsection{BabyAI Results}

\textbf{Factored Observations}:
The largest performance differences in Table~\ref{babyaimodelperformance} stem from the choice of factored versus flat or native image formats.
Notably, WMG with factored images achieved sample efficiencies 10x greater (on Level 3) than CNN+GRU using the native 7x7 image format.
However, factored observations alone are not sufficient for sample efficiency: 
WMG utilized factored images much more effectively than a GRU on Levels 2-5.
These large gains in sample efficiency support our hypothesis that Transformer-based models are particularly well suited for operating on set-based inputs like factored observations.

\textbf{Memos}:
Without factored observations, WMG-flat outperformed GRU-flat, suggesting that shortcut recurrence based on WMG's Memos compares favorably to the GRU's gated recurrence.
With the benefit of factored observations, the non-recurrent ablation of WMG (nr-WMG) performed slightly better than the full WMG on the simplest two levels.
But for the more challenging levels 3-5, Memos proved to be important for the best sample efficiency.

\textbf{Early vs Late instruction fusion}:
Interestingly, within our training limit of 6 million environment interactions, CNN+GRU was unable to learn to solve the levels (4 \& 5) where instructions carry important information. 
We suspect this is because the CNN processes just the image while the factored instruction is passed directly to the GRU, skipping the CNN.
By contrast, the baseline BabyAI agent uses FiLM layers~\citep{perez17} to integrate the processing of the image with the text instruction.
Both WMG and GRU models can process the image and instruction together in all levels of processing.
This early fusion appears to allow all WMG and GRU models to solve Level 4. 

In summary, the two WMG models with factored images were the only agents able to solve Level 5, and they learned to do so in approximately the same number of interactions that CNN+GRU required to solve Level 3.
These drastic differences in sample efficiency serve to highlight the potential gains that can be achieved by RL agents equipped to utilize factored observations.

While WMG's sample efficiencies dramatically improve upon the RL benchmarks published with the BabyAI domain \citep{Chevalier-2018}, often by two orders of magnitude (Table~\ref{babyailevels}), it's important to note that these sets of results are not directly comparable. Our experiments all used factored text instructions, and each model's hyperparameters were tuned for each level separately, while the BabyAI benchmark agent was trained on all levels using the single hyperparameter configuration provided in the BabyAI release. 

\begin{figure*}[htp]
    \centering
    \begin{subfigure}{0.23\textwidth}
    \includegraphics[width=\linewidth]{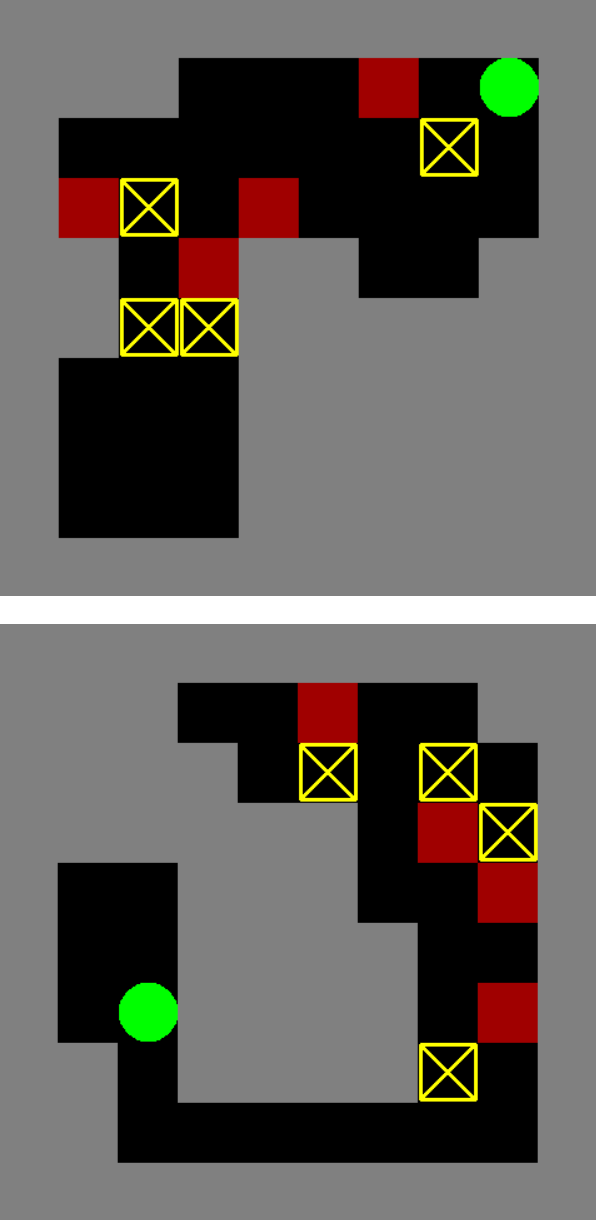}
    \end{subfigure}
    \begin{subfigure}{0.76\textwidth}
    \centering
    \includegraphics[width=0.96\linewidth]{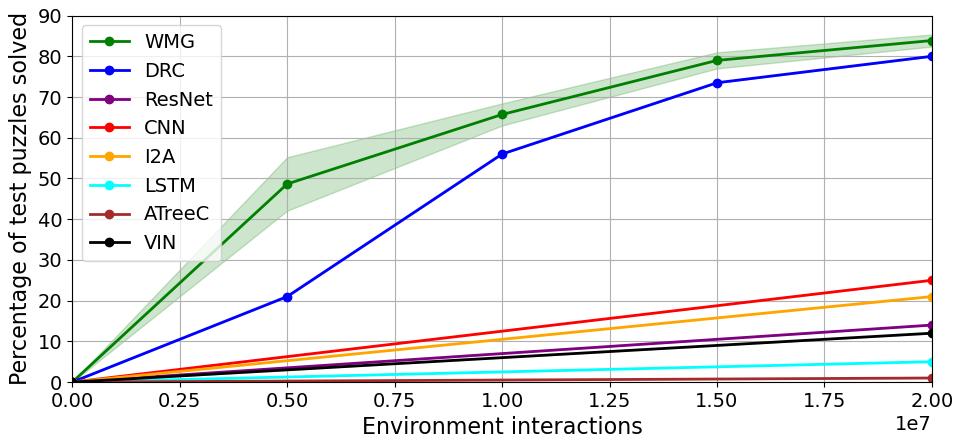}
    \end{subfigure}
    \vspace{-.5em}
    \caption{
    \textbf{Sokoban} puzzles (left) and \textbf{WMG results} (right)
    for 20 agents trained on the 900k-puzzle training set
    and evaluated on the 1000-puzzle test set from \citet{Sokoban},
    which reported the results for DRC and the other six agents shown here.
    Shading represents one standard deviation over the 20 independently trained WMG models.
    }
    \label{sokoban_perf_fig}
\end{figure*}

\subsubsection{Hyperparameter Sensitivity}
To evaluate WMG's sensitivity to hyperparameter selection, we applied the tuned hyperparameter settings from Level 4 to new training runs on all other levels. 
Figure~\ref{fig:hyperparam_sensitivity} shows moderate degradation in performance for all models. 
In particular, when the hyperparameter values tuned on Level 4 are used in Level 5 training runs, none of the models reach a 99\% solution rate within 1 million training steps, but WMG with factored observations reaches higher levels of performance than the other models.
Broadly, these results indicate that WMG is no more sensitive to hyperparameter settings than the baseline agents.

\subsection{Sokoban Environment}
As an independent test of WMG’s ability to perform complex reasoning over factored observations, we apply it to the Sokoban domain \citep{botea03}, a challenging puzzle that humans solve by forward planning. 
To successfully complete an episode, the agent (green circle in Figure~\ref{sokoban_perf_fig}, left) 
must push four yellow boxes onto the red targets within 120 timesteps.
Boxes may not be pushed into walls or other boxes, and cannot be pulled, so many moves render the puzzle unsolvable. 

We employ the training set, test set, and action space defined in \citet{Sokoban}.
Training episodes are sampled from 900k pre-generated puzzles.
The agent can move in each of the four directions (if not blocked) or stay in place.
The agent receives a reward of +1 for pushing a box onto a target, and -1 for pushing a box off of a target. 
Once all boxes are on targets, the agent receives a bonus reward of +2, and the episode ends. 

The observation space is factored as follows:
Each non-wall cell is represented by a factor containing 6 binary flags signifying whether the cell is a target, contains a box, and is bounded by a wall in each direction. 
The previous action and reward, plus information about the agent's currently occupied cell, are passed to the Core.
All other cells are mapped to separate Factors. 
Each Factor also receives two one-hot vectors specifying that cell's \textit{X} and \textit{Y} locations relative to the agent. 
Compared to image-based observations, our factored observations add egocentric information by encoding the relative positions of objects. 
The two spaces are otherwise isomorphic. 

We tuned hyperparameters on the training set (Appendix~\ref{appendix:hyperparameter_tuning}), 
then evaluated saved models from the corresponding twenty independent training runs on the 1000-puzzle test set. 
As shown in Fig \ref{sokoban_perf_fig} (right),
WMG quickly learned to solve most puzzles, and consistently outperformed DRC (Deep Repeated ConvLSTM) by \citet{Sokoban}.
WMG and DRC far exceeded the performance of the other six baseline agents, as reported in Table 2 of that work.
These results demonstrate that WMG can effectively use factored observations to solve difficult tasks that seem to rely on planning.

Videos of WMG tackling Sokoban puzzles are available at \url{https://tinyurl.com/vdz6gdd}.
To aid visualization of WMG's inner operations, many of the videos 
display white squares with areas proportional to the per-step attention probabilities applied by WMG's Core to all nodes in the preceding layer. These probabilities are summed over all attention heads, and over all layer pairs above the lowest. Attention applied to the single (in this task) Memo is represented by the white square in the upper-left cell. Our implementation of the Sokoban environment is based in part on that of \citet{SchraderSokoban2018}.

\section{Conclusion and future work}
We designed the \textit{Working Memory Graph} to investigate how self-attention can improve the memory and reasoning capabilities of RL agents. 
In contrast to previous models, WMG effectively leverages factored observations by encoding them into Factors and applying Transformer-style self-attention.
In order to reason over latent aspects of partially observable environments, WMG incorporates a novel form of recurrence using Memos to create multiple shortcut paths of self-attention. 

We compare WMG to gated RNN-based architectures (including state-of-the-art models like DRC) in three diverse environments featuring factored observations and complex reasoning over long-term dependencies. 
In these experiments, WMG outperforms competing models and demonstrates its ability to learn challenging tasks in a sample-efficient manner.

We stop short of claiming state-of-the-art performance on these domains, since factoring an observation space can alter the inherent difficulty of the task. 
Instead, our results demonstrate how a Transformer-based agent (WMG) can take advantage of factored observations to yield superior learning performance.

No model is without limitations. 
We conclude by outlining limitations of WMG as avenues for future work: 

\textbf{Flexible Memo lifetimes}:
In the current version of WMG, each new Memo automatically replaces the oldest. 
A more flexible and adaptive Memo-deletion scheme may improve WMG's ability to model latent aspects of the environment.
For instance, Memos that receive more attention than others may be the ones most worth keeping around for longer. 
Deleting a Memo only when its recently-received attention falls below a certain threshold would allow the number of Memos to fluctuate somewhat over time. 

\textbf{Graph edge content}:
As in the original Transformer, WMG applies input vectors to the nodes in its computation graph, but not to the edges between them. To better represent graph-structured data, \citet{1710.10903} contemplated incorporating edge-specific data into Transformer-based models. By harnessing the richer representational abilities of graphs over sets, 
a similar extension of WMG may allow it to better model complex relations among observed and latent factors in the environment.

\textbf{Memory vectors}:
Various forms of external memory have been proposed for RL agents \citep{graves2016hybrid, DBLP:journals/corr/abs-1907-09720}. 
Memory vectors retrieved from such stores could be passed to dedicated WMG \textit{memory} vectors, in addition to the Memos and Factors,
to further extend the range and flexibility of the agent's reasoning horizon.

\subsubsection*{Acknowledgments}
The authors wish to thank Alekh Agarwal and Xiaodong Liu for many valuable discussions, and Felipe Frujeri for creating the Sokoban videos.

\bibliography{icml2020_conference}
\bibliographystyle{icml2020_conference}

\clearpage

\begin{appendices}

\section{Pathfinding Task Details}
\label{appendix:pathfinding-environment-details}
The Pathfinding graph is constrained to be a polytree (singly connected, directed acyclic graph) at each step of an episode, as outlined in Algorithm~\ref{alg:pathfinding_dynamics}.

\begin{algorithm}[tbh]
   \caption{Pathfinding Episode Dynamics}
   \label{alg:pathfinding_dynamics}
\begin{algorithmic}
   \STATE {\bfseries Input:} pattern size $D$, max graph size $N$.
   \STATE Initialize $Graph = empty$.
   \STATE Add a node with random pattern $\in (-1, +1)^D$.
   \STATE $AddPattern = true$.
   \REPEAT
        \STATE {\bfseries Input:} agent $Action$.
        \STATE $Reward = 0$.
        \STATE $Done = false$.
        \IF {$AddPattern$ is $true$}
            \STATE // Construction step.
            \IF {$size(Graph) > 1$ and $Action == Target$}
    			\STATE $Reward = 1$.
    		\ENDIF
            \IF {$size(Graph) == N$}
    			\STATE $Done = true$.
    		\ELSE
    		    \STATE Choose random node $A$ from the Graph.
    		    \STATE Add node $B$ with random pattern $\in (-1, +1)^D$.
    		    \STATE Link $A$ and $B$ in a random direction.
    		    \STATE $Observation = A, B, 0$.
    		    \STATE $AddPattern = false$.
    		\ENDIF
    	\ELSE
    	    \STATE // Quiz step.
    	    \STATE Choose random $Target \in (true, false)$.
    	    \REPEAT
    	        \STATE Choose random nodes $X$ and $Y$.
    	        \STATE $PathExists =$ path exists from $X$ to $Y$.
    	    \UNTIL {$PathExists == Target$}
    	    \STATE $Observation = X, Y, 1$.
    	    \STATE $AddPattern = true$.
        \ENDIF
        \STATE {\bfseries Output:} $Reward, Observation, Done$
   \UNTIL{$Done$ is $true$}
\end{algorithmic}
\end{algorithm}

The hand-coded baseline agent is configured with a depth parameter $n$.
As new pattern pairs are revealed on graph-construction timesteps, this agent maintains a growing vector of all patterns seen, along with a growing matrix of directed path lengths from every observed pattern to every other. 
A path length of zero in this matrix indicates that no path exists from the first pattern to the second.
On each quiz step, the agent looks up from the matrix the path length $len$ for the ordered pair of patterns in the observation.
If $0 < len \leq n$, the agent chooses the \textit{yes} action. 
Otherwise, the agent chooses the \textit{no} action.

\section{Hyperparameter Tuning Procedures}
\label{appendix:hyperparameter_tuning}

\subsection{DGD}

In this work, all hyperparameters were tuned by a guided form of random search that we have named \textit{Distributed Grid Descent} (DGD). It is designed to address the challenges posed by large numbers of hyperparameters (10-20), and the high variance among independent training runs for the same hyperparameter configuration that is often observed in Deep RL experiments. DGD tackles these challenges by steering the random selection of configurations to be tested towards a robust basin, a fixed point in configuration space for which modification of any individual (discrete) hyperparameter setting by one step higher or lower results in worse performance in expectation over repeated runs. In addition, DGD is designed to run on multiple processes on potentially many machines with no central point of control. 

We define the following terms:

\textit{Tuning metric}:  A user-defined value calculated per training run for which higher is better, such as reward or success rate, or negative loss, etc. 

\textit{Run result}:  A completed run’s hyperparameter configuration and final tuning metric.

\textit{Run set}:  A single hyperparameter configuration, along with any available run results for that configuration.

\textit{Count(run set)}:  The number of runs in a run set.

\textit{Metric(run set)}:  The mean (or median) of the run metrics in a run set.

\textit{Neighborhood}:  A collection of run sets with configurations that differ by no more than one step in one setting from the configuration of a central run set in the neighborhood.

\textit{Count(neighborhood)}:  The maximum $Count$ of all run sets in the neighborhood.

The operation of each DGD worker process is described by Algorithm~\ref{alg:dgd_loop}.

\begin{algorithm}[tbh]
   \caption{Distributed Grid Descent}
   \label{alg:dgd_loop}
\begin{algorithmic}
   \STATE {\bfseries Input:} Set of hyperparameters $H$, each having a discrete, ordered set of possible values.
   \STATE {\bfseries Input:} Maximum number of training steps $N$ per run.
   \REPEAT
        \STATE {\bfseries Download} any new run results.
        \IF {no results so far}
            \STATE {\bfseries Choose} a random configuration $C$ from the grid defined by $H$.
        \ELSE
            \STATE {\bfseries Identify} the run set $S$ with the highest $Metric$.
            \STATE {\bfseries Initialize} neighborhood $B$ to contain only $S$.
            \STATE {\bfseries Expand} $B$ by adding all possible run sets whose configurations differ from that of $S$ by one step in exactly one hyperparameter setting.
            \STATE {\bfseries Calculate} a ceiling $M = Count(B) + 1$.
            \STATE {\bfseries Weight} each run set $x$ in $B$ by $M - Count(x)$.
            \STATE {\bfseries Sample} a random run set $S'$ from $B$ according to run set weights.
            \STATE {\bfseries Choose} the configuration $C$ from $S'$.
        \ENDIF
        \STATE {\bfseries Perform} one training run of $N$ steps using $C$.
        \STATE {\bfseries Calculate} the run’s $Metric$.
        \STATE {\bfseries Log} the run result to shared storage.
   \UNTIL terminated by user.
\end{algorithmic}
\end{algorithm}

Throughout the DGD search, the \textit{best} current run set is determined through Bayesian inference based on each run set's $Metric$ and $Count$, to filter out high-variance run sets having high average scores but relatively few runs. After the best posterior performance remains stable for some number of runs, the DGD search is terminated, and the best run set’s hyperparameter configuration is taken as the output of the search. To minimize the effects of local optima, the best run set can be chosen from a number of independent DGD searches.

\subsection{Application of DGD}

For the Pathfinding and BabyAI experiments, we ran five parallel DGD hyperparameter searches to convergence for each model, using the full number of training steps per run for the given experiment, and chose the best run set's hyperparameter configuration after convergence.

For WMG on Sokoban, we performed 60 DGD searches using training runs of 1.5M steps, with puzzle completion rate as the tuning metric. After those searches converged, we selected the best 25 configurations based on the training set results, and initiated 20 new training runs for each configuration. 
After training each agent for 10M environment interactions, we selected the single best hyperparameter configuration based on its performance on the training set. 
We then branched this configuration into 10 sets of runs with learning rate annealed every 100k steps using one of 10 separate values of gamma ranging from 0.6 to 0.98. 
After each agent was trained for a total of 20 million environment interactions, the annealing rate of 0.98 was found to perform the best on the training set. 
Finally, for this selected configuration's 20 independent agents, the models cached at 5M-step intervals were evaluated on 
the held-out test set of 1000 puzzles, producing the results shown in Figure~\ref{sokoban_perf_fig} (right).

All experiments were performed on Linux virtual machines in the Microsoft Azure cloud. The virtual machines featured Intel 2.6GHz Xeon E5 2667 v3 processors with 8 virtual CPUs, and no GPUs. 

\newpage

\onecolumn

\section{Supplemental Tables}
\label{appendix:appendix_tables}

\begin{table*}[htb]
\small
\caption{
Fixed settings and options used for all experiments, apart from the replicated BabyAI baselines in Table~\ref{babyailevels}.
}
\label{fixed_settings_table}
\begin{center}
\begin{tabular}{l | l}
\textbf{Settings and options} & \textbf{Values} \\
\hline
Dropout	&	None	\\
Learning rate schedule	&	Constant learning rate, except where noted	\\
Non-linearities	&	ReLU, tanh	\\
Parallel training workers	&	1	\\
Optimizer	&	Adam \citep{Adam}	\\
Parameter initialization, biases	&	0	\\
Parameter initialization, non-bias weights	&	Kaiming uniform \citep{KaimingUniform}	\\
Reward shaping	&	None	\\
Training algorithm	&	A3C \citep{Mnih16Asynchronous}	\\
Weight decay regularization	&	None	\\
\hline
\end{tabular}
\end{center}
\end{table*}

\begin{table*}[htb]
\small
\caption{
Hyperparameter values considered.
}
\label{all_hp_values_table}
\begin{center}
\begin{tabular}{l | l}
\hline
A3C $t_{max}$	&	1, 2, 3, 4, 6, 8, 10, 12, 16, 20, 24, 28, 32, 40, 48, 56, 64, 96, 120	\\
Actor-critic hidden layer size	&	64, 90, 128, 180, 256, 360, 512, 720, 1024, 1440, 2048, 2880, 4096, 5760	\\
Adam eps	&	1e-2, 1e-4, 1e-6, 1e-8, 1e-10, 1e-12	\\
CNN channel size 1 & 12, 16, 20 \\
CNN channel size 2 & 24, 32, 40 \\
CNN channel size 3 & 64, 128, 192 \\
Discount factor $\gamma$	&	0.5, 0.6, 0.7, 0.8, 0.9, 0.95, 0.98, 0.99, 0.995, 0.998	\\
Entropy term strength $\beta$	&	0.0, 0.002, 0.005, 0.01, 0.02, 0.05, 0.1, 0.2	\\
Gradient clipping threshold	&	2, 4, 8, 16, 32, 64, 128, 256, 512, 1024, 2048	\\
GRU observation embed size  & 128, 256, 512, 1024, 2048, 4096	\\
GRU size  & 64, 96, 128, 192, 256, 384, 512, 768, 1024	\\
Learning rate	&	4e-6, 6.3e-6, 1e-5, 1.6e-5, 2.5e-5, 4e-5, 6.3e-5, 1e-4, 1.6e-4, 2.5e-4, 4e-4	\\
Learning rate annealing $\gamma$ & 0.60, 0.64, 0.68, 0.72, 0.76, 0.80, 0.84, 0.88, 0.93, 0.98 \\
Reward on success (Sokoban) & 2, 5, 10, 15, 20 \\
Reward per step (Sokoban) & 0, -0.01, -0.02 \\
Reward scale factor	&	0.25, 0.5, 1, 2, 4, 8, 16, 32, 64, 128	\\
WMG attention head size	&	8, 12, 16, 24, 32, 48, 64, 90, 128, 180, 256, 360, 512	\\
WMG attention heads	&	1, 2, 3, 4, 6, 8, 10, 12, 16, 20	\\
WMG Memo size	&	32, 45, 64, 90, 128, 180, 256, 360, 512, 720, 1024, 1440, 2048	\\
WMG Memos	&	1, 2, 3, 4, 6, 8, 10, 12, 16, 20 	\\
WMG hidden layer size	&	6, 8, 12, 16, 24, 32, 48, 64, 96, 128, 192, 256, 384, 512	\\
WMG layers	&	1, 2, 3, 4, 5, 6, 7, 8, 9, 10, 11, 12, 13	\\
\hline
\end{tabular}
\end{center}
\end{table*}

For the Pathfinding domain, we verified in separate experiments that there is no difference in performance between setting the number of WMG Memos to $12$ (the maximum episode length) or $16$ (chosen by tuning). 
Since we introduce no penalty for model complexity, both DGD and random search would be expected to choose randomly between these possible values.

\begin{table*}[htb]
\small
\caption{
Tuned hyperparameter settings for Pathfinding experiments of 20M steps.
}
\label{pathfinding_hp_table}
\begin{center}
\begin{tabular}{l | l l l}
	&	\textbf{WMG}	&	\textbf{nr-WMG}	&	\textbf{GRU}	\\
\hline
Actor-critic hidden layer size	&	128	&	128	&	512	\\
A3C $t_{max}$	&	16	&	16	&	16	\\
Adam eps	&	1e-06	&	1e-08	&	1e-08	\\
Discount factor $\gamma$	&	0.5	&	0.6	&	0.5	\\
Entropy term strength $\beta$	&	0.01	&	0.005	&	0.02	\\
Gradient clipping threshold	&	16.0	&	16.0	&	4.	\\
GRU observation embedding size	&	-	&	-	&	256	\\
GRU size	&	-	&	-	&	384	\\
Learning rate	&	0.00016	&	0.00016	&	0.0001	\\
Reward scale factor	&	2.0	&	1.0	&	0.5	\\
WMG attention head size	&	12	&	16	&	-	\\
WMG attention heads	&	6	&	6	&	-	\\
WMG Memos	&	16	&	0	&	-	\\
WMG Memo size	&	128	&	-	&	-	\\
WMG hidden layer size	&	12	&	32	&	-	\\
WMG layers	&	4	&	4	&	-	\\
\hline
\end{tabular}
\end{center}
\end{table*}

\begin{table*}[htb]
\small
\caption{
Tuned hyperparameter settings for Pathfinding experiments of 1M steps.
}
\label{pathfinding_hp_table_2}
\begin{center}
\begin{tabular}{l | l l l}
	&	\textbf{WMG}	&	\textbf{nr-WMG}	&	\textbf{GRU}	\\
\hline
Actor-critic hidden layer size	&	2880	&	256	&	5760	\\
A3C $t_{max}$	&	16	&	16	&	16	\\
Adam eps	&	1e-04	&	1e-12	&	1e-06	\\
Discount factor $\gamma$	&	0.5	&	0.5	&	0.5	\\
Entropy term strength $\beta$	&	0.005	&	0.05	&	0.1	\\
Gradient clipping threshold	&	256.0	&	4.0	&	32.	\\
GRU observation embedding size	&	-	&	-	&	1024	\\
GRU size	&	-	&	-	&	256	\\
Learning rate	&	0.00004	&	0.00016	&	0.0001	\\
Reward scale factor	&	4.0	&	2.0	&	2.0	\\
WMG attention head size	&	90	&	90	&	-	\\
WMG attention heads	&	4	&	1	&	-	\\
WMG Memos	&	6	&	0	&	-	\\
WMG Memo size	&	256	&	-	&	-	\\
WMG hidden layer size	&	8	&	16	&	-	\\
WMG layers	&	3	&	5	&	-	\\
\hline
\end{tabular}
\end{center}
\end{table*}

\begin{table*}[htb]
\small
\caption{
Tuned hyperparameter settings for BabyAI Level 1 - GoToObj.
}
\label{babyai_l1_hp_table}
\begin{center}
\begin{tabular}{l | r r r r r r}
	&	\textbf{WMG}	&	\textbf{nr-WMG}	&	\textbf{GRU}	&	\textbf{WMG}	&	\textbf{GRU}	&	\textbf{CNN+GRU}	\\
	&	\textbf{factored}	&	\textbf{factored}	&	\textbf{factored}	&	\textbf{flat}	&	\textbf{flat}	&	\textbf{native 7x7x3}	\\
\hline
Actor-critic hidden layer size	&	2048	&	4096	&	4096	&	4096	&	2048	&	512	\\
A3C $t_{max}$	&	1	&	1	&	6	&	16	&	4	&	6	\\
Adam eps	&	0.0001	&	1e-08	&	1e-08	&	1e-10	&	0.0001	&	1e-10	\\
CNN hidden channel size 1	&	-	&	-	&	-	&	-	&	-	&	16	\\
CNN hidden channel size 2	&	-	&	-	&	-	&	-	&	-	&	40	\\
CNN hidden channel size 3	&	-	&	-	&	-	&	-	&	-	&	192	\\
Discount factor $\gamma$	&	0.98	&	0.9	&	0.7	&	0.6	&	0.9	&	0.8	\\
Entropy term strength $\beta$	&	0.002	&	0.05	&	0.01	&	0.005	&	0.02	&	0.02	\\
Gradient clipping threshold	&	256.0	&	1024.0	&	512.0	&	512.0	&	128.0	&	128.0	\\
GRU observation embed size	&	-	&	-	&	1024	&	-	&	512	&	512	\\
GRU size	&	-	&	-	&	96	&	-	&	512	&	96	\\
Learning rate	&	0.0001	&	4e-05	&	0.0004	&	0.0001	&	0.0001	&	0.0004	\\
Reward scale factor	&	4.0	&	32.0	&	32.0	&	8.0	&	32.0	&	8.0	\\
WMG attention head size	&	24	&	16	&	-	&	16	&	-	&	-	\\
WMG attention heads	&	4	&	10	&	-	&	12	&	-	&	-	\\
WMG Memos	&	1	&	0	&	-	&	1	&	-	&	-	\\
WMG Memo size	&	64	&	-	&	-	&	256	&	-	&	-	\\
WMG hidden layer size	&	64	&	64	&	-	&	32	&	-	&	-	\\
WMG layers	&	4	&	4	&	-	&	1	&	-	&	-	\\
\hline
\end{tabular}
\end{center}
\end{table*}

\begin{table*}[htb]
\small
\caption{
Tuned hyperparameter settings for BabyAI Level 2 - GoToRedBallGrey.
}
\label{babyai_l2_hp_table}
\begin{center}
\begin{tabular}{l | r r r r r r}
	&	\textbf{WMG}	&	\textbf{nr-WMG}	&	\textbf{GRU}	&	\textbf{WMG}	&	\textbf{GRU}	&	\textbf{CNN+GRU}	\\
	&	\textbf{factored}	&	\textbf{factored}	&	\textbf{factored}	&	\textbf{flat}	&	\textbf{flat}	&	\textbf{native 7x7x3}	\\
\hline
Actor-critic hidden layer size	&	4096	&	2048	&	4096	&	4096	&	4096	&	64	\\
A3C $t_{max}$	&	8	&	6	&	16	&	1	&	1	&	1	\\
Adam eps	&	1e-06	&	1e-08	&	1e-10	&	1e-10	&	1e-06	&	0.0001	\\
CNN hidden channel size 1	&	-	&	-	&	-	&	-	&	-	&	12	\\
CNN hidden channel size 2	&	-	&	-	&	-	&	-	&	-	&	24	\\
CNN hidden channel size 3	&	-	&	-	&	-	&	-	&	-	&	192	\\
Discount factor $\gamma$	&	0.8	&	0.9	&	0.8	&	0.9	&	0.9	&	0.95	\\
Entropy term strength $\beta$	&	0.01	&	0.02	&	0.01	&	0.005	&	0.005	&	0.02	\\
Gradient clipping threshold	&	1024.0	&	512.0	&	1024.0	&	128.0	&	64.0	&	64.0	\\
GRU observation embed size	&	-	&	-	&	4096	&	-	&	2048	&	256	\\
GRU size	&	-	&	-	&	96	&	-	&	512	&	64	\\
Learning rate	&	0.0001	&	0.00025	&	0.0001	&	2.5e-05	&	2.5e-05	&	0.0004	\\
Reward scale factor	&	8.0	&	4.0	&	4.0	&	4.0	&	4.0	&	2.0	\\
WMG attention head size	&	64	&	48	&	-	&	64	&	-	&	-	\\
WMG attention heads	&	4	&	1	&	-	&	3	&	-	&	-	\\
WMG Memos	&	1	&	0	&	-	&	8	&	-	&	-	\\
WMG Memo size	&	32	&	-	&	-	&	64	&	-	&	-	\\
WMG hidden layer size	&	16	&	24	&	-	&	384	&	-	&	-	\\
WMG layers	&	3	&	3	&	-	&	1	&	-	&	-	\\
\hline
\end{tabular}
\end{center}
\end{table*}

\begin{table*}[htp]
\small
\caption{
Tuned hyperparameter settings for BabyAI Level 3 - GoToRedBall.
}
\label{babyai_l3_hp_table}
\begin{center}
\begin{tabular}{l | r r r r r r}
	&	\textbf{WMG}	&	\textbf{nr-WMG}	&	\textbf{GRU}	&	\textbf{WMG}	&	\textbf{GRU}	&	\textbf{CNN+GRU}	\\
	&	\textbf{factored}	&	\textbf{factored}	&	\textbf{factored}	&	\textbf{flat}	&	\textbf{flat}	&	\textbf{native 7x7x3}	\\
\hline
Actor-critic hidden layer size	&	4096	&	2048	&	4096	&	4096	&	4096	&	4096	\\
A3C $t_{max}$	&	1	&	2	&	3	&	1	&	2	&	3	\\
Adam eps	&	1e-12	&	0.0001	&	1e-06	&	0.0001	&	1e-06	&	0.01	\\
CNN hidden channel size 1	&	-	&	-	&	-	&	-	&	-	&	12	\\
CNN hidden channel size 2	&	-	&	-	&	-	&	-	&	-	&	40	\\
CNN hidden channel size 3	&	-	&	-	&	-	&	-	&	-	&	192	\\
Discount factor $\gamma$	&	0.95	&	0.9	&	0.9	&	0.9	&	0.9	&	0.9	\\
Entropy term strength $\beta$	&	0.1	&	0.05	&	0.1	&	0.05	&	0.02	&	0.05	\\
Gradient clipping threshold	&	128.0	&	128.0	&	128.0	&	128.0	&	32.0	&	32.0	\\
GRU observation embed size	&	-	&	-	&	2048	&	-	&	4096	&	256	\\
GRU size	&	-	&	-	&	192	&	-	&	512	&	64	\\
Learning rate	&	2.5e-05	&	6.3e-05	&	6.3e-05	&	2.5e-05	&	2.5e-05	&	0.0004	\\
Reward scale factor	&	8.0	&	4.0	&	8.0	&	8.0	&	4.0	&	4.0	\\
WMG attention head size	&	128	&	32	&	-	&	24	&	-	&	-	\\
WMG attention heads	&	2	&	8	&	-	&	12	&	-	&	-	\\
WMG Memos	&	2	&	0	&	-	&	16	&	-	&	-	\\
WMG Memo size	&	128	&	-	&	-	&	256	&	-	&	-	\\
WMG hidden layer size	&	64	&	32	&	-	&	128	&	-	&	-	\\
WMG layers	&	4	&	4	&	-	&	1	&	-	&	-	\\
\hline
\end{tabular}
\end{center}
\end{table*}

\begin{table*}[htp]
\small
\caption{
Tuned hyperparameter settings for BabyAI Level 4 - GoToLocal.
}
\label{babyai_l4_hp_table}
\begin{center}
\begin{tabular}{l | r r r r r}
	&	\textbf{WMG}	&	\textbf{nr-WMG}	&	\textbf{GRU}	&	\textbf{WMG}	&	\textbf{GRU}	\\
	&	\textbf{factored}	&	\textbf{factored}	&	\textbf{factored}	&	\textbf{flat}	&	\textbf{flat}	\\
\hline
Actor-critic hidden layer size	&	2048	&	2048	&	1024	&	512	&	4096	\\
A3C $t_{max}$	&	6	&	3	&	3	&	6	&	4	\\
Adam eps	&	1e-12	&	0.01	&	1e-06	&	1e-08	&	1e-12	\\
Discount factor $\gamma$	&	0.5	&	0.6	&	0.95	&	0.5	&	0.9	\\
Entropy term strength $\beta$	&	0.1	&	0.1	&	0.1	&	0.02	&	0.02	\\
Gradient clipping threshold	&	512.0	&	512.0	&	256.0	&	256.0	&	512.0	\\
GRU observation embed size	&	-	&	-	&	1024	&	-	&	512	\\
GRU size	&	-	&	-	&	128	&	-	&	96	\\
Learning rate	&	6.3e-05	&	0.0001	&	4e-05	&	2.5e-05	&	4e-05	\\
Reward scale factor	&	32.0	&	16.0	&	8.0	&	16.0	&	2.0	\\
WMG attention head size	&	128	&	64	&	-	&	24	&	-	\\
WMG attention heads	&	2	&	4	&	-	&	16	&	-	\\
WMG Memos	&	8	&	0	&	-	&	16	&	-	\\
WMG Memo size	&	32	&	-	&	-	&	64	&	-	\\
WMG hidden layer size	&	32	&	48	&	-	&	16	&	-	\\
WMG layers	&	4	&	3	&	-	&	2	&	-	\\
\hline
\end{tabular}
\end{center}
\end{table*}

\begin{table*}[htb]
\small
\caption{
Tuned hyperparameter settings for BabyAI Level 5 - PickupLoc.
}
\label{babyai_l5_hp_table}
\begin{center}
\begin{tabular}{l | r r}
	&	\textbf{WMG}	&	\textbf{nr-WMG}	\\
	&	\textbf{factored}	&	\textbf{factored}	\\
\hline
Actor-critic hidden layer size	&	512	&	2048	\\
A3C $t_{max}$	&	12	&	12	\\
Adam eps	&	1e-10	&	1e-10	\\
Discount factor $\gamma$	&	0.7	&	0.8	\\
Entropy term strength $\beta$	&	0.02	&	0.05	\\
Gradient clipping threshold	&	512.0	&	512.0	\\
Learning rate	&	0.0001	&	6.3e-05	\\
Reward scale factor	&	8.0	&	8.0	\\
WMG attention head size	&	24	&	48	\\
WMG attention heads	&	10	&	6	\\
WMG Memos	&	8	&	0	\\
WMG Memo size	&	32	&	-	\\
WMG hidden layer size	&	128	&	96	\\
WMG layers	&	2	&	2	\\
\hline
\end{tabular}
\end{center}
\end{table*}

\begin{table*}[htb]
\small
\caption{
Tuned hyperparameter values on Sokoban.
The resulting model contained 4,508,182 trainable parameters.
}
\label{sokoban_hp_values_table}
\begin{center}
\begin{tabular}{l | r r}
\hline
Reward per step & 0 \\
Reward on success & 2 \\
Actor-critic hidden layer size	&	2880	\\
A3C $t_{max}$	&	4	\\
Adam eps	&	1e-10	\\
Discount factor $\gamma$	&	0.995	\\
Entropy term strength $\beta$	&	0.02	\\
Gradient clipping threshold	&	512.0	\\
Learning rate	&	1.6e-5	\\
Learning rate annealing $\gamma$ & 0.98 \\
Reward scale factor	&	4	\\
WMG attention head size	&	32	\\
WMG attention heads	&	8	\\
WMG Memos	&	1	\\
WMG Memo size	&	2048	\\
WMG hidden layer size	&	8	\\
WMG layers	&	10	\\
\hline
\end{tabular}
\end{center}
\end{table*}


\begin{table*}[htb]
\small
\caption{Additional details for Pathfinding experiments of 20M steps.}
\label{pathfinding_detailed_results}
\begin{center}
\begin{tabular}{l | r r r}
Models \& algorithms & Final performance & Trainable parameters & Training speed \\ 
\hline
\textit{Depth-(n-1)} baseline & 100.0\% of reward \\
\textit{Depth-3} baseline & 99.7\% of reward \\
\textit{Depth-2} baseline & 97.6\% of reward \\
\textit{Depth-1} baseline & 86.9\% of reward \\
nr-WMG, full-history & 99.6\% of reward & 204,963 & 96 steps/sec \\
WMG & 99.6\% of reward & 132,507 & 91 steps/sec \\
GRU & 94.7\% of reward & 1,139,459 & 291 steps/sec \\
\hline
\end{tabular}
\end{center}
\end{table*}

\begin{table*}[htp]
\small
\caption{Number of trainable parameters, in thousands, for the BabyAI models in Table~\ref{babyaimodelperformance}.}
\label{babyaimodelsize}
\begin{center}
\begin{tabular}{l | r r r r r r r r}
& WMG & nr-WMG & GRU & WMG & GRU & CNN+GRU \\ 
BabyAI level & factored & factored & factored & flat & flat & native 7x7x3 \\ 
\hline
1 - GoToObj & 636 & 1,864 & 1,572 & 2,053 & 4,170 & 393 \\
2 - GoToRedBallGrey & 2,997 & 258 & 3,723 & 2,116 & 10,075 & 140 \\
3 - GoToRedBall & 3,418 & 2,217 & 3,749 & 3,229 & 15,126 & 709 \\
4 - GoToLocal & 2,235 & 1,960 & 1,137 & 2,022 & 1,479 & ----- \\
5 - PickupLoc & 879 & 2,007 & ----- & ----- & ----- & ----- \\
\hline
\end{tabular}
\end{center}
\end{table*}

\begin{table*}[htp]
\small
\caption{Training steps per second on a fixed machine, for the BabyAI models in Table~\ref{babyaimodelperformance}.}
\label{babyaimodelspeed}
\begin{center}
\begin{tabular}{l | r r r r r r r r}
& WMG & nr-WMG & GRU & WMG & GRU & CNN+GRU \\ 
BabyAI level & factored & factored & factored & flat & flat & native 7x7x3 \\ 
\hline
1 - GoToObj & 38 & 28 & 146 & 111 & 86 & 149 \\
2 - GoToRedBallGrey & 58 & 113 & 147 & 35 & 18 & 88 \\
3 - GoToRedBall & 18 & 32 & 78 & 25 & 20 & 87 \\
4 - GoToLocal & 44 & 48 & 132 & 54 & 134 & ----- \\
5 - PickupLoc & 81 & 84 & ----- & ----- & ----- & ----- \\
\hline
\end{tabular}
\end{center}
\end{table*}

\begin{table*}[htp]
\small
\caption{
BabyAI baseline agent sample efficiencies, defined as the amount of training (in either episodes or environment interaction steps) required for the agent to solve 99\% of random episodes within 64 steps.
The published results are the means of the min \& max RL sample efficiencies reported in Table 3 of \citet{Chevalier-2018}.
We obtained the replicated results, which are the medians over 10 training runs, using the code and default hyperparameter settings from the open source release of the BabyAI baseline agent. 
We report these sample efficiencies in terms of both episodes and environment interactions.
All numbers are in thousands.
}
\label{babyailevels}
\begin{center}
\begin{tabular}{l | l | r r r r r}
& & Published & Replicated & Replicated \\ 
BabyAI level & Instruction template & (\textit{episodes}) & (\textit{episodes}) & (env interactions) \\ 
\hline
1 - GoToObj & GO TO $\langle$color$\rangle$ $\langle$object$\rangle$ & ----- & \textit{19} & 333 \\
2 - GoToRedBallGrey & GO TO RED BALL & \textit{17} & \textit{16} & 282 \\
3 - GoToRedBall & GO TO RED BALL & \textit{297} & \textit{283} & 3,674 \\
4 - GoToLocal & GO TO $\langle$color$\rangle$ $\langle$object$\rangle$ & \textit{1008} & \textit{1,064} & 16,422 \\
5 - PickupLoc & PICK UP $\langle$color$\rangle$ $\langle$object$\rangle$ $\langle$location$\rangle$ & \textit{1,545} & \textit{1,557} & 25,574 \\
\hline
\end{tabular}
\end{center}
\end{table*}

\end{appendices}

\end{document}